\documentclass[letterpaper, 10 pt, conference]{./support/ieeeconf}
\usepackage{times}
\usepackage[pdftex]{graphicx}
\usepackage{subfigure}
\usepackage{amsmath,amssymb,amsopn,amstext,amsfonts}
\usepackage{cancel}
\usepackage[space]{cite}
\usepackage{pdfsync}
\usepackage{balance}
\usepackage{color}
\usepackage{mathtools}
\usepackage[ruled,vlined]{algorithm2e}
\usepackage{bm}

\usepackage{diagbox}
\usepackage{float}
\usepackage{epstopdf}
\usepackage{pifont}
\usepackage{fixltx2e}
\usepackage{amsmath}
\usepackage{multirow}
\usepackage{booktabs}
\usepackage{url}
\usepackage{svg}
\usepackage{threeparttable}
\usepackage{graphicx}
\usepackage{caption}
\usepackage{lipsum}
\usepackage[linkcolor=black,citecolor=black,urlcolor=black,colorlinks=true]{hyperref}

\newcommand{\tp}{^{\mathrm{T}}}

\newcommand{\df}[1]{\mathrm{d}{#1}}
\newcommand{\rbrac}[1]{({#1})}

\newcommand{\norm}[1]{\Vert{#1}\Vert}

\graphicspath{{./figures/}}
\DeclareGraphicsExtensions{.png,.jpg,.eps,.pdf}
\IEEEoverridecommandlockouts
\makeatletter
\def\endthebibliography{%
  \def\@noitemerr{\@latex@warning{Empty `thebibliography' environment}}%
  \endlist
}

\title{\LARGE \bf Real-Time Trajectory Planning for Aerial Perching}
\author{Jialin Ji, Tiankai Yang, Chao Xu, and Fei Gao
\thanks{This work was supported by National Natural Science Foundation of China under Grant 62003299 and 62088101.}
\thanks{The State Key Laboratory of Industrial Control Technology, College of Control Science and Engineering, Zhejiang University, Hangzhou 310027, China, and Huzhou Institute, Zhejiang University, Huzhou 313000, China.}
\thanks{Email: {\tt\small\{jlji, fgaoaa\}@zju.edu.cn}}
\thanks{Corresponding Author: Fei Gao.}
}

\begin{document}

\makeatletter
\let\@oldmaketitle\@maketitle
\renewcommand{\@maketitle}{\@oldmaketitle
	\captionsetup{type=figure}
	\includegraphics[width=\linewidth]
	{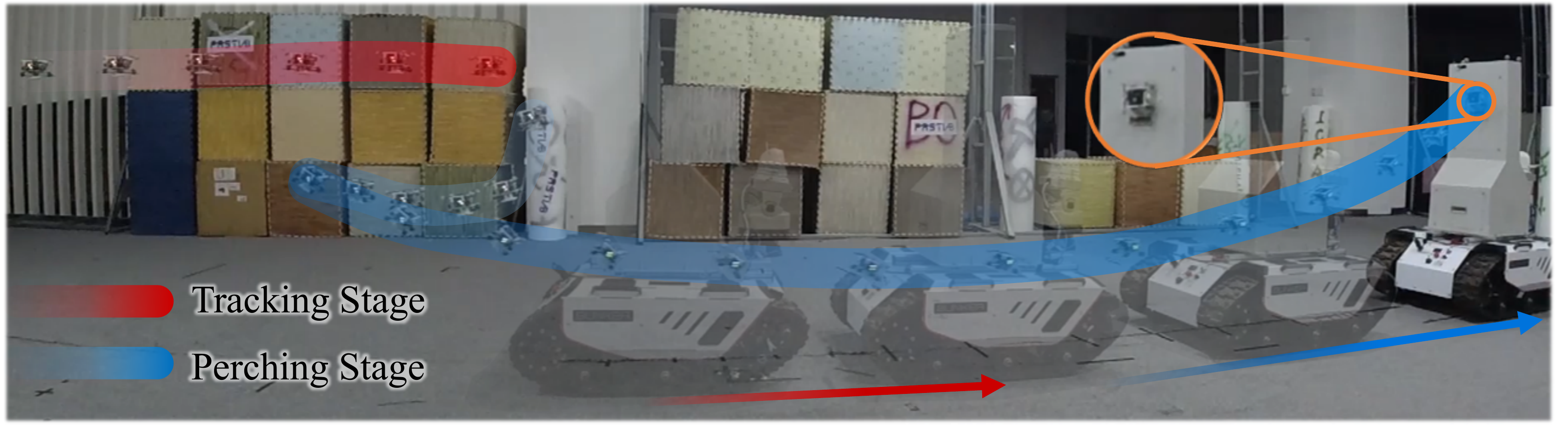}
	\label{fig:introduction}
	\captionof{figure}{
	Aerial perching on the vertical surface of a moving platform. 
	The drone first tracks the ground vehicle (red) and then plans and carries out the perching trajectory (blue) when triggered. 
	Since our drone has low thrust-to-weight ratio (1.7) and is close to the ground, our planner generates a swing-shaped trajectory guaranteeing both safety and dynamic feasibility.
}
}
\makeatother

\maketitle

\begin{abstract}
  This paper presents a novel trajectory planning method for aerial perching. 
  Compared with the existing work, the terminal states and the trajectory durations can be adjusted adaptively, instead of being determined in advance. Furthermore, our planner is able to minimize the tangential relative speed on the premise of safety and dynamic feasibility. This feature is especially notable on micro aerial robots with low maneuverability or scenarios where the space is not enough.
  Moreover, we design a flexible transformation strategy to eliminate terminal constraints along with reducing optimization variables.
  Besides, we take precise $\mathrm{SE(3)}$ motion planning into account to ensure that the drone would not touch the landing platform until the last moment. 
  The proposed method is validated onboard by a palm-sized micro aerial robot with quite limited thrust and moment (thrust-to-weight ratio 1.7) perching on a mobile inclined surface. 
  Sufficient experimental results show that our planner generates an optimal trajectory within 20ms, and replans with warm start in 2ms.
\end{abstract}

\section{Introduction}
Autonomous aerial robots are widely applied in many fields, with the advantages of lightweight, great sensitivity and high maneuverability.
However, their limited payload and flight time dramatically hinders the leverage of the above strengths.
Some researchers investigate landing and perching on variant surfaces to reduce energy consumption while the drones conduct monitoring, sampling or recharging. 
These work successfully directs the robots towards the targets and adjust their postures during the maneuver. 

However, these methods generally set the desired terminal poses and velocities of the robots, which may lead to dynamic infeasibility when there's no enough space. 
Besides, they cannot handle complex constraints and achieve high computational efficiency simultaneously. 
Generally, there are still three intractable requirements for planning an optimal perching trajectory:
\begin{enumerate}
  \item Terminal states adjustment: To get greater freedom of optimization, both the flight duration and the terminal state of the robot should be adjusted adaptively, rather than being determined in advance.
  \item Collision avoidance: The drone is not allowed to touch the landing platform until the last moment, which requires SE(3) motion planning.
  \item High-frequency replanning: Due to changes of the landing point caused by perception disturbance or ego-motion, frequent replanning is necessary.
\end{enumerate}

In this paper, we propose an effective joint optimization approach \footnote{\url{https://github.com/ZJU-FAST-Lab/Fast-Perching}} that enables generating a spatio-temporal optimal trajectory satisfying all the above requirements.
The method would significantly enhance the capability of perching, especially for perching on mobile platforms, such as ground vehicles or flying carriers.
For a perfect perching flight, the relative tangential velocity to the surface at the landing moment should be zero to avoid scratching the surface. 
Most existing perching systems set a fixed desired terminal velocity according to their variant perching mechanism: zero for grippers\cite{chi2014optimized, thomas2015visual, popek2018autonomous, hang2019perching, roderick2021bird, paneque2022perception}, a large normal velocity for suction \cite{tsukagoshi2015aerial, kessens2016versatile} and adhesives\cite{mellinger2012trajectory, daler2013perching, hawkes2013dynamic, kalantari2015autonomous, thomas2016aggressive, mao2021aggressive}.
Nevertheless, such constraint requires the drone to dive in advance and is prone to collisions with the ground when there is no enough space. 
This phenomenon is more common for micro UAVs with limited maneuverability. 
To solve this problem, we design a flexible terminal transformation to minimize the tangential relative speed on the premise of safety and dynamic feasibility.
As for the collision with the mobile target, most existing methods\cite{vlantis2015quadrotor} detect the intersection between a bounding volume, which encloses the aerial robot, and the infinite plane defined by the landing platform. 
However, such an approximation makes the generated trajectory too conservative. Instead,  we model the mobile platform as a disc and constrain the robot on the landing side of the platform only if the robot falls into its projection. 
A numerical technique is designed to transform such a complex mixed-integer nonlinear problem into a much simpler formulation.

We summarize our contributions as follows:

\begin{enumerate}
  \item A highly-versatile optimal trajectory generation method is proposed for perching on moving inclined surfaces.
  \item A general analytical formulation is designed for dynamic collision avoidance between the robot and the mobile platform, which fully considers SE(3) motion planning.
  \item A flexible terminal transformation is introduced so that complex constraints can be eliminated and the terminal state can be adjusted adaptively to guarantee both safety and dynamic feasibility.
  \item Sufficient simulations and real-world experiments validate the performance of our implementation. Planning can be carried out within 20ms, and replanning with warm start costs less than 2ms.
\end{enumerate}

\section{Related Work}

Most prior work \cite{mellinger2012trajectory, daler2013perching, hawkes2013dynamic, kalantari2015autonomous, tsukagoshi2015aerial, thomas2016aggressive, kessens2016versatile, mao2021aggressive} studies the problem of perching on vertical walls. They usually design particular perching mechanisms such as suction or adhesive grippers. Then the robots can attach to smooth surfaces like glass or tiles, leveraging the specified terminal normal speed.
There is also some work \cite{chi2014optimized, thomas2015visual, popek2018autonomous, hang2019perching, roderick2021bird, paneque2022perception} focusing on the problem of perching on cylindrical objects or cables. In these methods, robots approach the targets gradually while hovering and then use claws or other similar gripers to perch on the objects.

Many existing methods \cite{mellinger2012trajectory, thomas2016aggressive, mao2021aggressive} parameterize trajectories with piece-wise polynomials and carry out planning in flat-output space. This approach constructs a more solvable problem but is difficult to set nonlinear constraints.
Thomas et al. \cite{thomas2016aggressive} propose a planning and control strategy for quadrotors as well as a customized dry adhesive gripper to achieve the necessary conditions for perching on smooth surfaces. 
This method fully considers actuator constraints and formulates a quadratic programming (QP) using a series of linear approximations. 
This approach cannot adjust the flight time of the trajectory, and the linearization is oversimplified.
Mao et al. \cite{mao2021aggressive} also formulate a QP to constrain the terminal state and velocity bound. 
They increase the trajectories’ time iteratively and recursively solve the QP until the thrust constraint is satisfied, which runs much faster than solving the nonlinear programming (NLP) directly. 
However, this method cannot be used for limiting the angular velocity or other complex nonlinear constraints. 
Besides, simply extending the flight time does not necessarily guarantee dynamic feasibility for aerial perching.

Other methods \cite{vlantis2015quadrotor, paneque2022perception} formulate a constrained discrete-time non-linear MPC to satisfy complex constraints but cost much more computation time.
Vlantis et al. \cite{vlantis2015quadrotor} first study the problem of landing a quadrotor on an inclined moving platform. 
The method designs the aerial robot’s control inputs such that it initially approaches the platform, while maintaining it within the camera’s field of view, and finally lands on it.
However, such a complex problem is computed on a ground station, and the robot only runs a low-level controller onboard. 
Moreover, the slope of the inclined surface is quite small, which is not so difficult to land on.
Paneque et al. \cite{paneque2022perception} formulate a discrete-time multiple-shooting NLP problem for perching on powerlines, which computes perception-aware, collision-free, and dynamically-feasible maneuvers to guide the robot to the desired final state. 
Notably, the generated maneuvers consider both the perching and the posterior recovery trajectories. 
Nevertheless, this method also suffers excessive computation time and cannot adjust the terminal state smartly. 

\section{Modeling and Nomenclature}

In this paper, we use the simplified dynamics proposed by \cite{mueller2015computationally} for a quadrotor, whose configuration is defined by its translation $p \in \mathbb R^3$ and rotation $R \in \mathrm{SO}(3)$. Translational motion depends on the gravitational acceleration $\bar g$ as well as the thrust $\tilde f$. Rotational motion takes the body rate $\omega \in \mathbb R^3$ as input. The simplified model is written as 
\begin{equation}
  \begin{cases}
  \tau = \tilde f R \mathbf e_3 / m, \\
    \ddot{p} = \tau - \bar{g} \mathbf e_3, \\
    \dot{R} = R \hat \omega,
  \end{cases}
\end{equation}
where $\tau$ denotes the net thrust, $\mathbf e_i$ is the $i$-th column of $\mathbf I_3$ and $\hat \cdot$ is the skew-symmetric matrix form of the vector cross product.
\begin{figure}[t]
	\begin{center}
		\includegraphics[width=1.0\columnwidth]{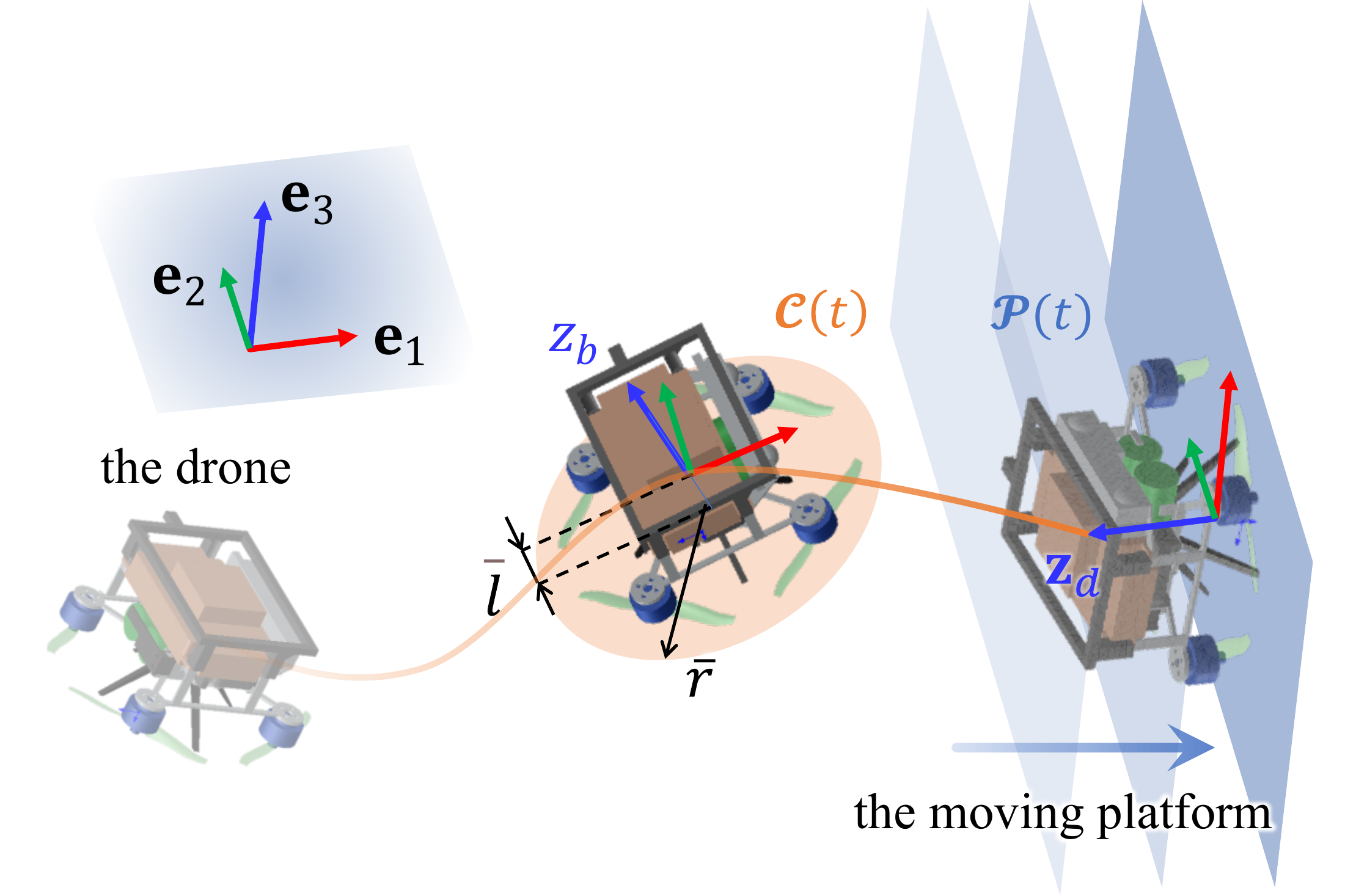}
	\end{center}
	\caption{
		\label{fig:model}
		Modeling overview and symbols definition.
	}
\vspace{-0.5cm}
\end{figure}
Moreover, we model the underside of a symmetric quadrotor as a disc, shown in Fig. \ref{fig:model}, denoted by
\begin{equation}
  \label{eq:C(t)}
  \mathcal C(t) = \left\{ x = R(t) \mathbf B \bar r u + c(t) ~\Big|~ \norm{u} \leq 1, u \in \mathbb R_2 \right\},
\end{equation}
where $\mathbf B = \left(\mathbf e_1, \mathbf e_2\right) \in \mathbb R_{3\times2}$ and $\bar r$ denotes the radius of the disc. We use $\bar l$ to denote the length of the drone's centroid to the bottom. Thus the center of the disc $c(t) = p(t) - \bar l z_b(t)$.

We assume that the future position of the landing platform is estimated as $\varrho(t)$ and the normal vector of the desired perching surface is denoted by $\mathbf z_d$. Therefore the perching plane of the moving platform is written as
\begin{equation}\\
  \label{eq:P(t)}
  \mathcal P(t) = \left\{ \mathbf a^T x \leq b(t) ~\Big|~ x \in \mathbb R_3 \right\},
\end{equation}
where $\mathbf a = - \mathbf z_d, b(t) = \mathbf a^T\varrho(t)$.

Exploiting the differential flatness property, we conduct optimization in the flat-output space of multicopters.
In this paper, we adopt $\mathfrak{T}_{\mathrm{MINCO}}$ \cite{wang2021geometrically}, a minimum control effort polynomial trajectory class. 
An $s$-order $\mathfrak{T}_{\mathrm{MINCO}}$ is in fact a 2$s$-order polynomial spline with constant boundary conditions. 
It provides a linear-complexity smooth map from intermediate points $q$ and a time allocation $T$ to the coefficients of splines $c$, denoted by $c = \mathcal{M}(q, T)$. Such a spline is the unique control effort minimizer of an s-integrator passing $q$. 
Besides, given any cost function $\mathcal F(c, T)$, the corresponding function $  \mathcal J(q, T) = \mathcal F(\mathcal M(q, T), T)$.
$\mathfrak{T}_{\mathrm{MINCO}}$ gives a linear-complexity way to compute $\partial \mathcal J / \partial q$ and $\partial \mathcal J / \partial T$ from corresponding $\partial \mathcal F / \partial c$ and $\partial \mathcal F / \partial c$. After that, a high-level optimizer is able to optimize the objective efficiently.

\section{Planning for Aggressive Perching}

\subsection{Problem Formulation}

After estimating the future motion of the moving target $\varrho(t)$, we expect a smooth, collision free and dynamically feasible trajectory $p(t)$, whose terminal speed and orientation coincide with the ones of the inclined surface.
Concluding the above requirements of optimal perching gives the following problem:
\begin{subequations}
  \begin{align}
    \min_{p(t), T} & \label{eq:cost function}~\mathcal J_o = \int_{0}^{T} {\norm{p^{(s)}(t)}^2} \df{t} + \rho T,                                                      \\
    s.t.~          & \label{eq:T>0} ~T > 0, \\
    & \label{eq:init pos}~p^{[s-1]}(0)=\mathbf p_o,                                                                                                    \\
                   & \label{eq:final pos}~p(T)=\varrho(T) - \bar l \mathbf z_d,                                                                                       \\
                   & \label{eq:perching zb}~z_b(T) = \mathbf z_d,                                                                                                     \\
                   & \label{eq:velocity limitation}~\norm{p^{(1)}(t)}\leq v_{max}, ~\forall t\in[0,T],                                                                \\
                   & \label{eq:body rate limitation}~\norm{\omega(t)}\leq \omega_{max}, ~\forall t\in[0,T],                                                           \\
                   & \label{eq:thrust limitation}~\tau_{min} \leq \norm{\tau(t)} \leq \tau_{max}, ~\forall t\in[0,T],                                                 \\
                   & \label{eq:ground}~\mathbf e_3^T p(t) \geq z_{min}, ~\forall t\in[0,T],                                                                           \\
                   & \label{eq:perching relative velocity}~p^{(1)}(T) = \varrho^{(1)}(T) ~\text{(in the ideal case)},                                                 \\
                   & \label{eq:perching collision}~\mathcal C(t) \in \mathcal P(t) \text{~if~} \norm{p(t)-\varrho(t)} \leq \bar d,               
  \end{align}
\end{subequations}
where cost function Eq. \ref{eq:cost function} trades off the smoothness and aggressiveness. 
Actuator constraints include speed Eq. \ref{eq:velocity limitation}, body rate Eq. \ref{eq:body rate limitation} and thrust Eq. \ref{eq:thrust limitation} limitations. 
Eq. \ref{eq:ground} and Eq. \ref{eq:perching collision} are safety constraints for ground and the landing platform, where $\bar d$ denotes the size of the perching surface.
Boundary conditions involve initial state Eq. \ref{eq:init pos}, terminal position Eq. \ref{eq:final pos}, perching pose Eq. \ref{eq:perching zb} and terminal relative velocity Eq. \ref{eq:perching relative velocity}.  
Noted that the modification of Eq. \ref{eq:perching relative velocity} under non-ideal condition will be discussed later in \ref{sec:Flexible Terminal Transformation}.

We use $\mathfrak{T}_{\mathrm{MINCO}}$ of $s=4$ and $N$-piece for minimum snap and enough freedom of optimization. Then the gradients $\partial{\mathcal J_o}/\partial c$ and $\partial{\mathcal J_o}/\partial T$ can be evaluated as
\begin{subequations}
  \begin{align}
    \frac{\partial{\mathcal J_o}}{\partial c_i} = & 2\left( \int_0^{T_i} \beta^{(3)}(t)\beta^{(3)}(t)\tp \df{t} \right) c_i, \\
    \frac{\partial{\mathcal J_o}}{\partial T_i} = & c_i\tp\beta^{(3)}(T_i)\beta^{(3)}(T_i)\tp c_i + \rho,
  \end{align}
\end{subequations}
where $\beta(t)=\rbrac{1,t,\dots,t^N}\tp$ is the natural basis.

\subsection{Actuator Constraints}
\begin{figure}[t]
	\begin{center}
		\includegraphics[width=1.0\columnwidth]{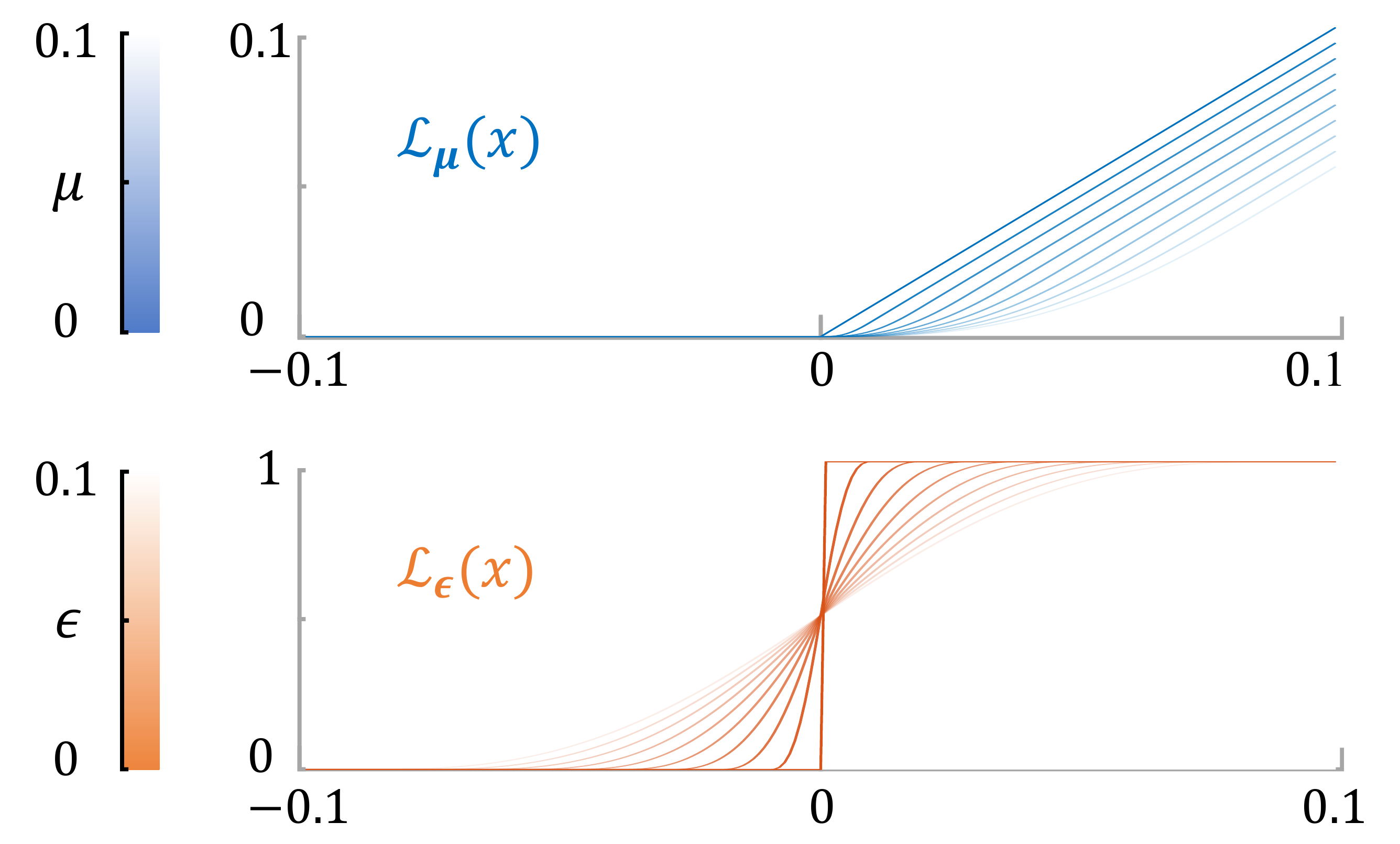}
	\end{center}
	\caption{
		\label{fig:L_fun}
		As $\mu$ and $\epsilon$ approach 0, $C^2$-Smoothing of the exact penalty $\mathcal L_\mu[\cdot]$ and smoothed logistic function $\mathcal L_\epsilon[\cdot]$ vary to the ideal ones.
	}
\vspace{-0.5cm}
\end{figure}
Firstly, the net thrust $\tau(t) = p^{(2)}(t) + \bar{g} \mathbf e_3$ is bounded as Eq. \ref{eq:thrust limitation}. Inspired by \cite{wang2021robust}, it can be constrained by constructing such a penalty function
\begin{align}
  \mathcal G_{\tau} (t) = \mathcal L_\mu [\norm{\tau(t)}^2 - \tau_{max}^2 ] + \mathcal L_\mu [ \tau_{min}^2 - \norm{\tau(t)}^2 ],
\end{align}
where $\mathcal L_\mu[\cdot]$is an $C^2$-smoothing of the exact penalty, shown in Fig. \ref{fig:L_fun}, which is denoted by
\begin{align}
  \mathcal L_\mu[x] = 
  \begin{cases}
    0,                    & x \leq 0,       \\
    (\mu - x/2)(x/\mu)^3, & 0 < x \leq \mu, \\
    x - \mu/2,            & x > \mu.
  \end{cases}
\end{align}

Secondly, the limitation of body rate $\omega = R^T(t)\dot R(t)$ can also be constrained by a similar penalty function
\begin{align}
  \mathcal G_{\omega} (t) & = \mathcal L_\mu \left[ \norm{\omega(t)}^2 - \tau_{max}^2 \right].
\end{align}
To avoid singularity and simplify calculation, we enforce this constraint by using
Hopf fibration \cite{watterson2020control}, thus
\begin{subequations}
  \begin{align}
     & z_b(t) = R(t)\mathbf e_3 = \tau(t) / \norm{\tau(t)}_2,                  \\
     & \label{eq:fdn}\dot z_b(t) = f_{\cal DN}\left[\tau(t)\right] p^{(3)}(t), \\
     & \omega_1^2 + \omega_2^2 = \norm{\dot{z_b}}, 
  \end{align}
\end{subequations}
where $z_b$ is aligned with the direction of thrust according to the simplified dynamics \cite{mueller2015computationally}, and $\omega_i$ denotes the $i$-th component of angular velocity $\omega$. Function $f_{\cal DN}[\cdot]$ in Eq. \ref{eq:fdn} is given by
\begin{align}
  f_{\cal DN}[x] = \left(\mathbf{I}_3 - \frac{xx^T}{x^Tx}\right) / \norm{x}_2.
\end{align}
Since the angle yaw $\psi$ changes little while perching, $\omega_3$ can be ignored.

Finally, the penalty of velocity limitation Eq. \ref{eq:velocity limitation} can also be written as a penalty
\begin{align}
  \mathcal G_{v} (t) & = \mathcal L_\mu \left[ \norm{p^{(1)}(t)}^2 - v_{max}^2 \right].
\end{align}

\subsection{Collision Avoidance}
The constraint avoiding the collision with ground Eq. \ref{eq:ground} can be transformed as such a penalty function just like previous constraints
\begin{align}
  \mathcal G_g(t) = \mathcal L_\mu \left[z^2_{min} - \norm{\mathbf e_3^T p(t)}^2 \right].
\end{align}

Given the description of $\mathcal C(t)$ (Eq. \ref{eq:C(t)}) and $\mathcal P(t)$ (Eq. \ref{eq:P(t)}), the constraint $\mathcal C(t) \in \mathcal P(t)$ in Eq. \ref{eq:perching collision} is equivalent to
\begin{subequations}
  \begin{align}
    \mathbf a^T \left(R(t) \mathbf B \bar r u + c(t) \right) - b(t) \leq 0,                   \\
    \sup_{\norm{u} \leq 1} \mathbf a^T \left(R(t) \mathbf B \bar r u \right) + \mathbf a^Tc(t) - b(t) \leq 0, \\
    \label{eq:F(t)}
    \bar r \norm{\mathbf B^T R(t)^T \mathbf a} + \mathbf a^T c(t) - b(t) \leq 0.
  \end{align}
\end{subequations}
According to differential flatness with Hopf fibration \cite{watterson2020control}, if we denote $z_b = [a, b, c]^T \in S^2$, we can write the following unit quaternion $q_{a b c}$ satisfying $R\left(q_{a b c}\right) \mathbf e_{3} = z_b$,
\begin{align}
  q_{a b c}=\frac{1}{\sqrt{2(1+c)}}\left(\begin{array}{c}1+c \\-b \\a \\0\end{array}\right).
\end{align}
The robot's rotation $R$ without yaw can be obtained, thus
\begin{equation}
  \mathbf{B}^{T} R^T=
  \begin{pmatrix}
1-\frac{a^{2}}{1+c} & -\frac{a b}{1+c} & -a \\ -\frac{a b}{1+c} & 1-\frac{b^{2}}{1+c} & -b
  \end{pmatrix}
\end{equation}
However, the constraint Eq. \ref{eq:F(t)} should only be activated when $\norm{p(t) - \varrho(t)} \leq \bar d$, that is, the drone is close to the platform.
This constitutes a mixed-integer nonlinear programming problem.
Here we design a smoothed logistic function $\mathcal L_\epsilon[\cdot]$ to incorporate the integer variable into our NLP, which is denoted by
\begin{align}
\mathcal L_\epsilon[x] = 
\begin{cases}
0,                                     & x \leq -\epsilon,     \\
\frac{1}{2\epsilon^4}(x+\epsilon)^3(\epsilon-x), & -\epsilon < x \leq 0, \\
\frac{1}{2\epsilon^4}(x-\epsilon)^3(\epsilon+x) + 1, & 0 < x \leq \epsilon,  \\
1,                                     & x > \epsilon,
\end{cases}
\end{align}
where $\epsilon$ is a tunable positive parameter. Then the penalty function of Eq. \ref{eq:perching collision} can be written as
\begin{align}
	\mathcal G_c(t) &= \mathcal L_\mu \left[\mathcal F_1(t) \right] \cdot \mathcal L_\epsilon \left[\mathcal F_2(t) \right],
\end{align}
where $\mathcal F_1(t)$ and $\mathcal F_2(t)$ are denoted by
\begin{subequations}
	\begin{align}
	\mathcal F_1(t) &= \bar r \norm{\mathbf B^T R(t)^T \mathbf a} + \mathbf a^T c(t) - b(t), \\
	\mathcal F_2(t) &= \norm{p(t) - \varrho(t)}^2 - {\bar d}^2.
	\end{align}
\end{subequations}


\subsection{Flexible Terminal Transformation}
\label{sec:Flexible Terminal Transformation}

\begin{figure}[t]
  \begin{center}
    \includegraphics[width=1.0\columnwidth]{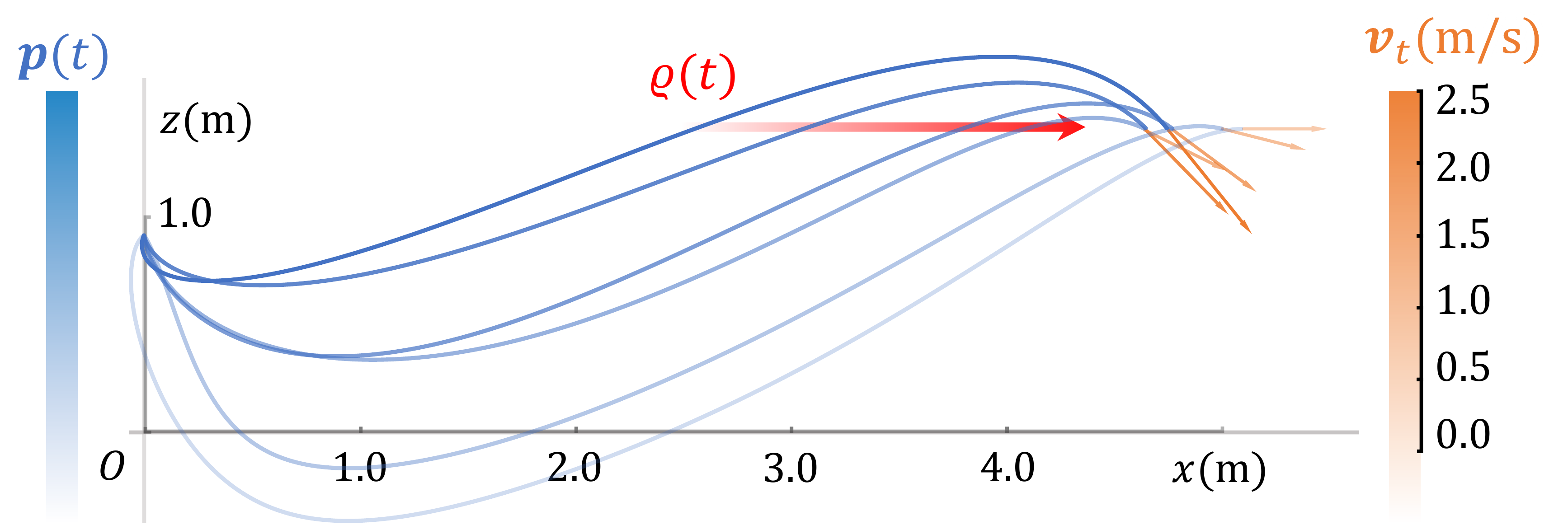}
  \end{center}
  \caption{
    \label{fig:vt}
    The optimized trajectories with different terminal vertical speeds $v_t$ under the limitation of dynamic feasibility.
    The ones with small $v_t$ collide the ground. The parameters are set as below: the height of landing position $1.5 \mathrm m$, $\tau_{max} = 17 \mathrm {m/s^2}$, $\tau_{min} = 5 \mathrm {m/s^2}$, and $\omega_{max} = 3 \mathrm{rad/s}$.
  }
\vspace{-0.5cm}
\end{figure}

Since we are planning trajectories in the flat-output space, the terminal state should be determined from $p^{[s-1]}(T)$. 
Therefore, except for the terminal position Eq. \ref{eq:final pos}, terminal velocity, acceleration and jerk should also be optimized or determined.
In the ideal case constraint Eq. \ref{eq:perching relative velocity} should be satisfied. 
Nevertheless, such hard constraint may lead to conflict with the collision avoidance constraint Eq. \ref{eq:ground} when there is no enough space. 
As shown in Fig. \ref{fig:vt}, under the limitation of rotor thrust and angular rate, the optimized trajectory with zero vertical speed will collide the ground. 

Therefore, we set the following transformation
\begin{align}
p^{(1)}(T) = \varrho^{(1)}(T) - \bar v_n \mathbf z_d + \mathbf V v_t,
\end{align}
where $\mathbf V = \left(\mathbf v_1, \mathbf v_2\right) \in \mathbb R_{3\times2}$ and $\mathbf v_1, \mathbf v_2$ are two orthogonal unit vectors perpendicular to $\mathbf z_b$. The desired normal relative speed $\bar v_n$ depends on perching mechanics and the tangential relative velocity $v_t \in \mathbb R^2$ becomes the new optimal variable instead of $p^{(1)}(T)$. Meanwhile, we minimize the tangential relative speed by introducing regulation term
\begin{align}
	\mathcal J_{t} = \norm{v_t}^2.
\end{align}

The constraint of terminal pose Eq. \ref{eq:perching zb} is equivalent to
\begin{align}
\tau(T) / \norm{\tau(T)}_2 = \mathbf z_d.
\end{align}
Considering this constraint and the thrust limit for terminal state, we design such a transformation
\begin{align}
	p^{(2)}(T) = \left(\tau_{m} + \tau_{r} \cdot \sin(\tau_{f})\right) \cdot \mathbf z_d - \bar g e_3,
\end{align}
where $\tau_{m} = (\tau_{max} + \tau_{min})/2$ and $\tau_{r} = (\tau_{max} - \tau_{min})/2$. Thus constraints Eq. \ref{eq:perching zb} and Eq. \ref{eq:thrust limitation} are both eliminated by replacing $p^{(2)}(T)$ with a new optimal variable $\tau_f \in \mathbb R$.

As for terminal jerk, since it is related to the terminal angular rate according to Eq. \ref{eq:fdn}, we should set $p^{(3)}(T) = \mathbf 0$ to make the relative angular velocity of the robot small when it touches the platform.

\subsection{Other Implementation Details}

All the above constraints written as penalty functions should be satisfied along the whole trajectory. 
Such infinite constraints cannot be solved directly. 
Inspired by \cite{wang2021geometrically}, we can transform them into finite constraints by using integral of these penalties.
For reasonable approximation, the time integral penalty $\mathcal J_\star $ with gradient can be easily derived by
\begin{subequations}
	\begin{align}
	\label{eq:PieceTimeIntegralPenalty}
	\mathcal I^\star_i=                                             & \frac{T_i}{\kappa}\sum_{j=0}^{\kappa_i}\bar{\omega}_j\mathcal{G}_\star \rbrac{\frac{j}{\kappa} T_i },      
	\mathcal J_\star = \sum_{i=1}^{M} \mathcal I^\star_i,                                                                                                                                                    \\
	\frac{\partial \mathcal J_\star}{\partial c_i} =              & \frac{\partial \mathcal I^\star_i}{\partial \mathcal G_\star} \frac{\partial \mathcal G_\star}{\partial c_i},
	~\frac{\partial \mathcal J_\star}{\partial T_i} =              \frac{\mathcal I^\star_i}{T_i} + \frac{j}{\kappa} \frac{\partial \mathcal I^\star_i}{\partial \mathcal G_\star}\frac{\partial \mathcal G_\star}{\partial t},
	\end{align}
\end{subequations}
where integer $\kappa$ controls the relative resolution of quadrature.  $\rbrac{\bar{\omega}_0,\bar{\omega}_1,\dots,\bar{\omega}_{\kappa_i-1},\bar{\omega}_{\kappa_i}}=\rbrac{1/2,1,\cdots,1,1/2}$ are the quadrature coefficients following the trapezoidal rule \cite{press2007numerical}. $i = (1, 2, \cdots, N)$ denotes the $i$-th piece and $j = (1, 2, \cdots, \kappa)$.

Summarizing the above strategies, we transform the original problem into an unconstrained nonlinear optimization problem, the cost function of which is given by
\begin{align}
	\mathcal J = \mathcal J_o + \sum  w_\star \mathcal J_\star, 
	~\star = \{\tau, \omega, v, g, c, t\},
\end{align}
where $w_\star$ are weight coefficients for each costs.

For optimization efficiency and uniform distribution of time allocation, we set duration of each pieces $T_i$ to be equal, and use transformation $T = e^{T'}$ to eliminate the constraint Eq. \ref{eq:T>0}. Thus the optimization variables are now
\begin{align}
	T' \in \mathbb R, q \in \mathbb R^{3\times N}, v_t \in \mathbb R^2, \tau_f \in \mathbb R.
\end{align}
We set initial value of $v_t$ to $\mathbf 0$, $\tau_f$ to $(\tau_{max} + \tau_{min})/2$ and $T'$ to $0$. A boundary value problem is solved to obtain the initial guess of $q$. After obtaining all the gradients of the optimization variables using the above approaches, the problem is then efficiently solved by the L-BFGS \cite{liu1989limited}.

\section{Perching Experiments}

\begin{figure}[t]
	\begin{center}
		\includegraphics[width=0.8\columnwidth]{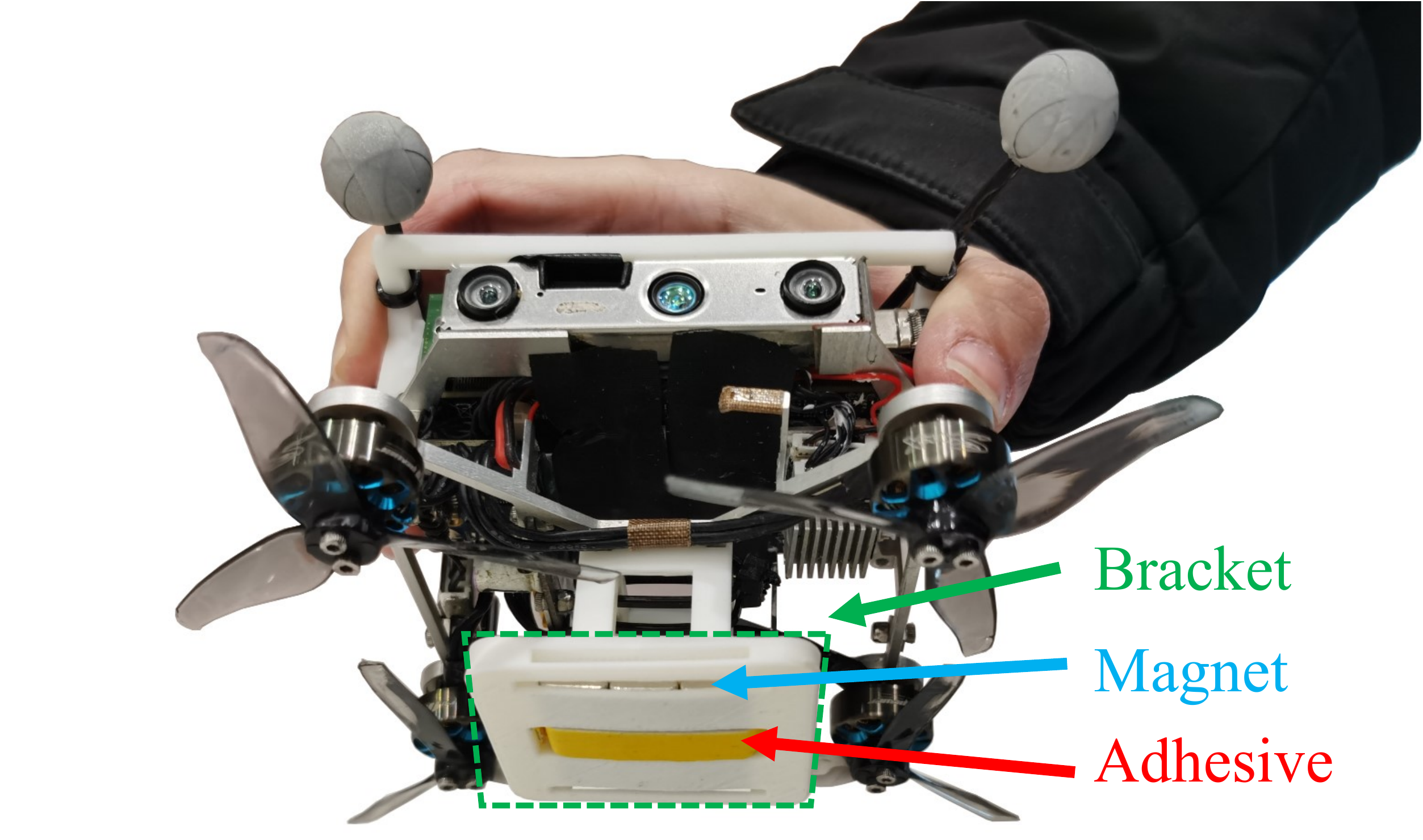}
	\end{center}
	\caption{
		\label{fig:drone}
		The Quadrotor platform and the perching mechanism used in our experiments.
	}
\end{figure}

Since our planner is suitable for most types of active/passive perching devices, we design an improvised perching mechanism for the experiments, shown in Fig. \ref{fig:drone}. 
This device is composed of a magnet that provides positive pressure to any iron surface and an adhesive providing sufficient friction. 
The quadrotor platform weights 0.36kg and has a thrust-to-weight ratio of 1.7. 
An NVIDIA Jetson Xavier NX is used as the main onboard computer and the state estimates of the quadrotor are given by an EKF of the pose from VICON and the IMU data from a PX4 Autopilot. 
We adopt the controller using Hopf fibration\cite{watterson2020control} to avoid singularities and align the attitude calculation of planning and control.

\begin{figure}[t]
	\begin{center}
		\includegraphics[width=1.0\columnwidth]{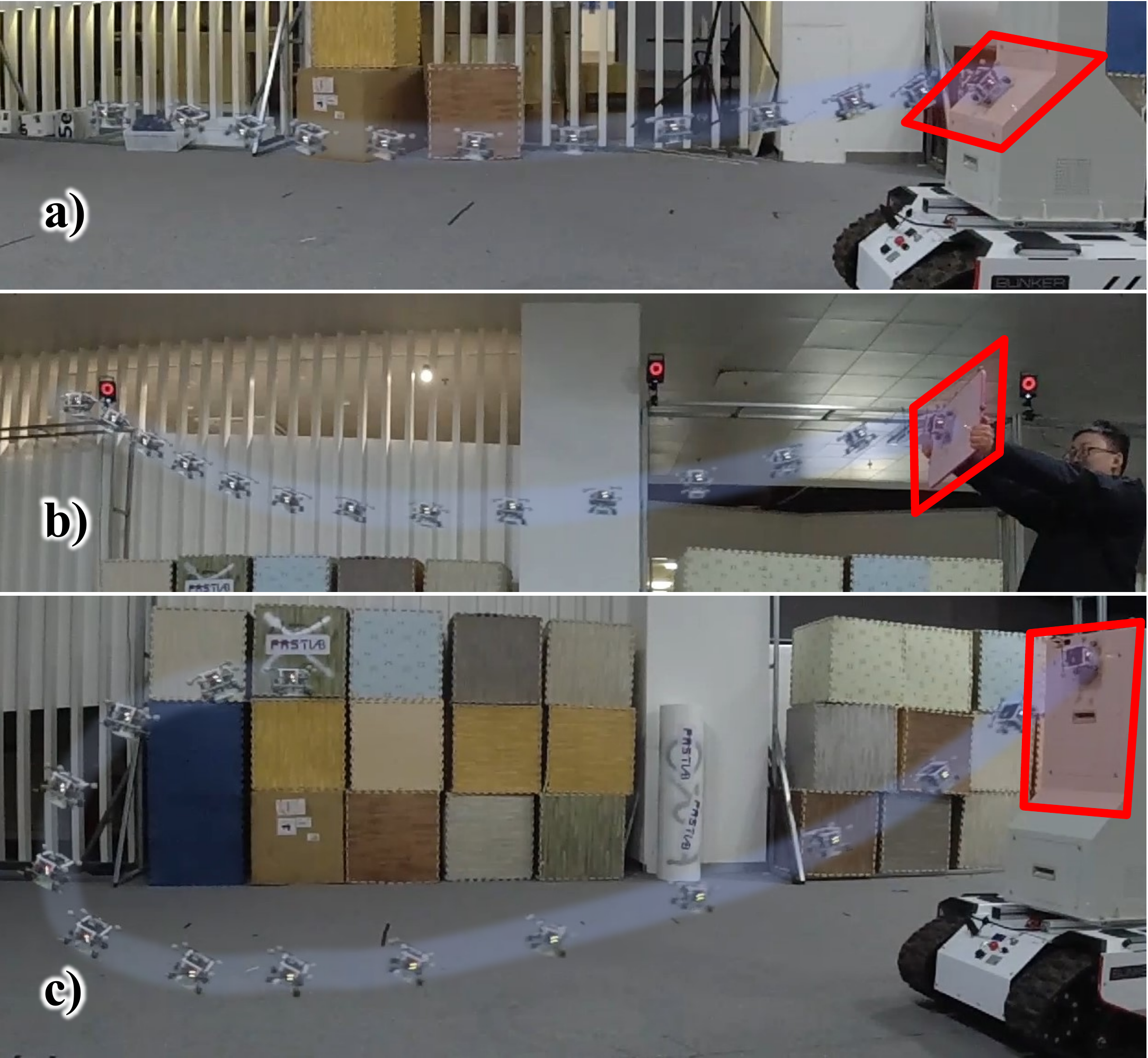}
	\end{center}
	\caption{
		\label{fig:three_static}
		Perching experiments of different inclined surfaces.
	}
\vspace{-1cm}
\end{figure}

We first conduct several experiments for perching on different static inclined surfaces, shown in Fig. \ref{fig:three_static}. 
Considering the errors of control and estimation, we set the desired terminal norm velocity $\bar v_n = 0.3 \mathrm{m/s}$ to ensure the magnet of the drone can be attached to the iron surfaces at the end of the trajectory. 
Other constraint parameters are set as
$v_{max} = 6 \mathrm{m/s}$, $\tau_{min} = 5 \mathrm{m/s^2}$, $\tau_{max} = 15 \mathrm{m/s^2}$, $\omega_{max} = 3 \mathrm{rad/s}$. 
As we can see, the drone is able to plan variant trajectories for different perching tasks: a) A $45^\circ$ inclined surface $3\mathrm  m$ away; b) A whiteboard of an arbitrary orientation raised by people $3\mathrm  m$ away; c) A vertical surface $2.5\mathrm  m$ away. The robustness of our method can be fully validated from these experimental results.

\begin{figure}[ht]
	\begin{center}
		\includegraphics[width=1.0\columnwidth]{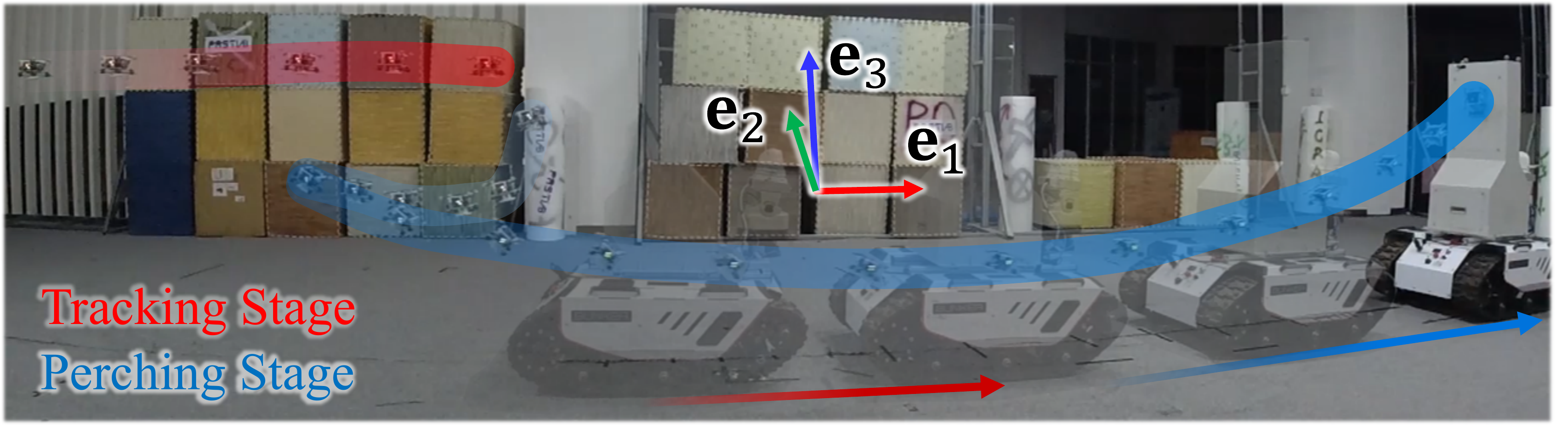}
	\end{center}
	\caption{
		\label{fig:moving90}
		Frame definitions of the dynamic experiment.
	}
\end{figure}

\begin{figure*}[t]
	\begin{center}
		\includegraphics[width=2.0\columnwidth]{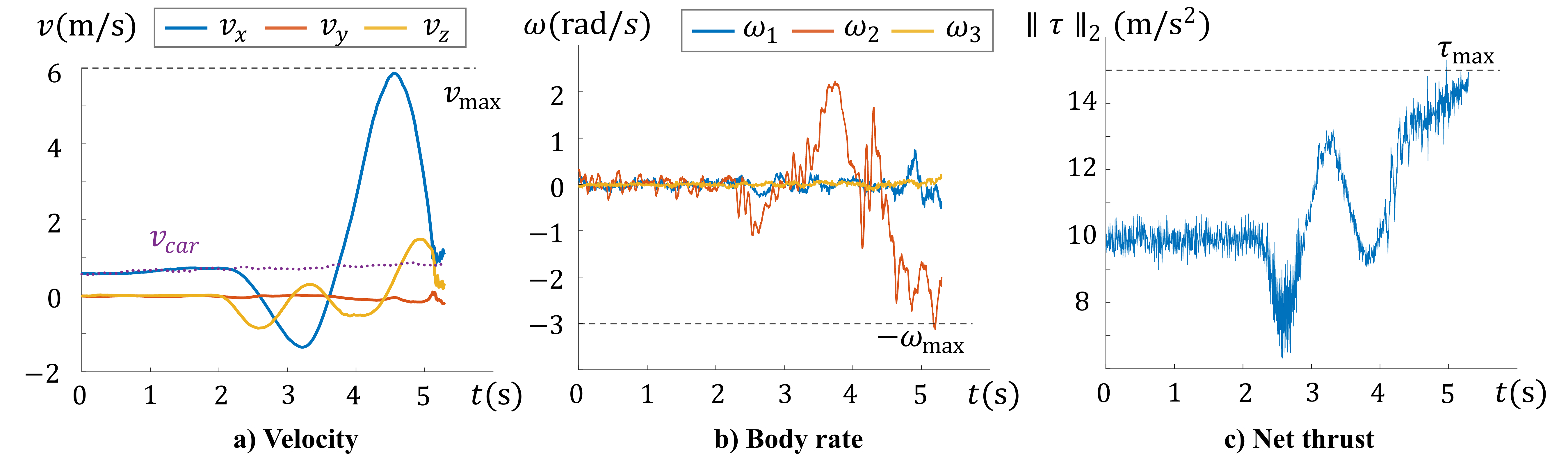}
	\end{center}
	\caption{
		\label{fig:experiment}
		a) Velocities of the drone to the ground vehicle.
		b) Body rate of the drone while perching maneuver.
		c) Net thrust calculated from imu.
	}
\end{figure*}

To validate the adaptiveness of our planner, we also conduct an experiment for perching on the inclined surface of a moving platform, shown in Fig. \ref{fig:moving90}. A ground robot is moving forward at the speed of $0.6\mathrm {m/s}$ and the aerial robot is following after it (red marker). While receiving the instruction, the drone plans a perching trajectory onboard and carries out replanning at 10hz. 
The horizontal distance between the drone and the ground vehicle is $2.3\mathrm m$ and the landing height $\mathbf z_t = 1.1\mathrm m$, which is quite close to the safety height $z_{min} = 0.4\mathrm m$. 
Moreover, the drone has quite limited thrust and moment.
Therefore, it's hard to successfully plan a smooth trajectory guaranteeing both safety and dynamic feasibility. 
We set the same constraint parameters as the previous experiments without any adjustments.
The swing-shaped trajectory generated by our planner satisfies the requirement. 
The detailed collected data is shown in Fig. \ref{fig:experiment}.
The desired terminal relative norm velocity $\bar v_n = 0.3 \mathrm{m/s}$ can be seen as $v_{rx}$ in Fig. \ref{fig:experiment} a). We also collect the data of imu and calculate the profile of body rate and net thrust, shown in Fig. \ref{fig:experiment} b) and Fig. \ref{fig:experiment} c). Both the angular velocity and the thrust are bounded and the drone finally lands smoothly on the vertical surface of the ground vehicle while moving.

\section{Evaluations}

We benchmark our proposed method with an open-source state-of-the-art perching method \cite{paneque2022perception}.
This work formulates a constrained discrete-time NLP problem and solves it using ForcesPRO\cite{FORCESPro}. Instead of using differential flatness like us, they model the inputs of the system as the desired constant thrust derivatives and control each rotor thrusts directly.
To compare the methods fairly, we set a loose constraint of each motor thrust and constrain the whole net thrust as the same as us so that the dynamics model of the work\cite{paneque2022perception} and ours are similar.
Since this method focuses on perception-awareness and perching on powerlines, which is certain targeted, we remove the perception, collision avoidance and other specific constraints. 
By the same token, we also fix the terminal state for our planner and remove the collision avoidance constraints. 
The initial and terminal positions of the robots are set to $(0, 0, 4.2)$ and $(4.0, 0, 4.25)$. Besides, the trajectories are required to be rest-to-rest. 
We set the same parameters of the dynamic constraints: $v_{max} = 6 \mathrm{m/s}$, $\tau_{min} = 5 \mathrm{m/s^2}$, $\tau_{max} = 17 \mathrm{m/s^2}$, $\omega_{max} = 3 \mathrm{rad/s}$. 

\begin{figure}[t]
	\begin{center}
		\includegraphics[width=0.8\columnwidth]{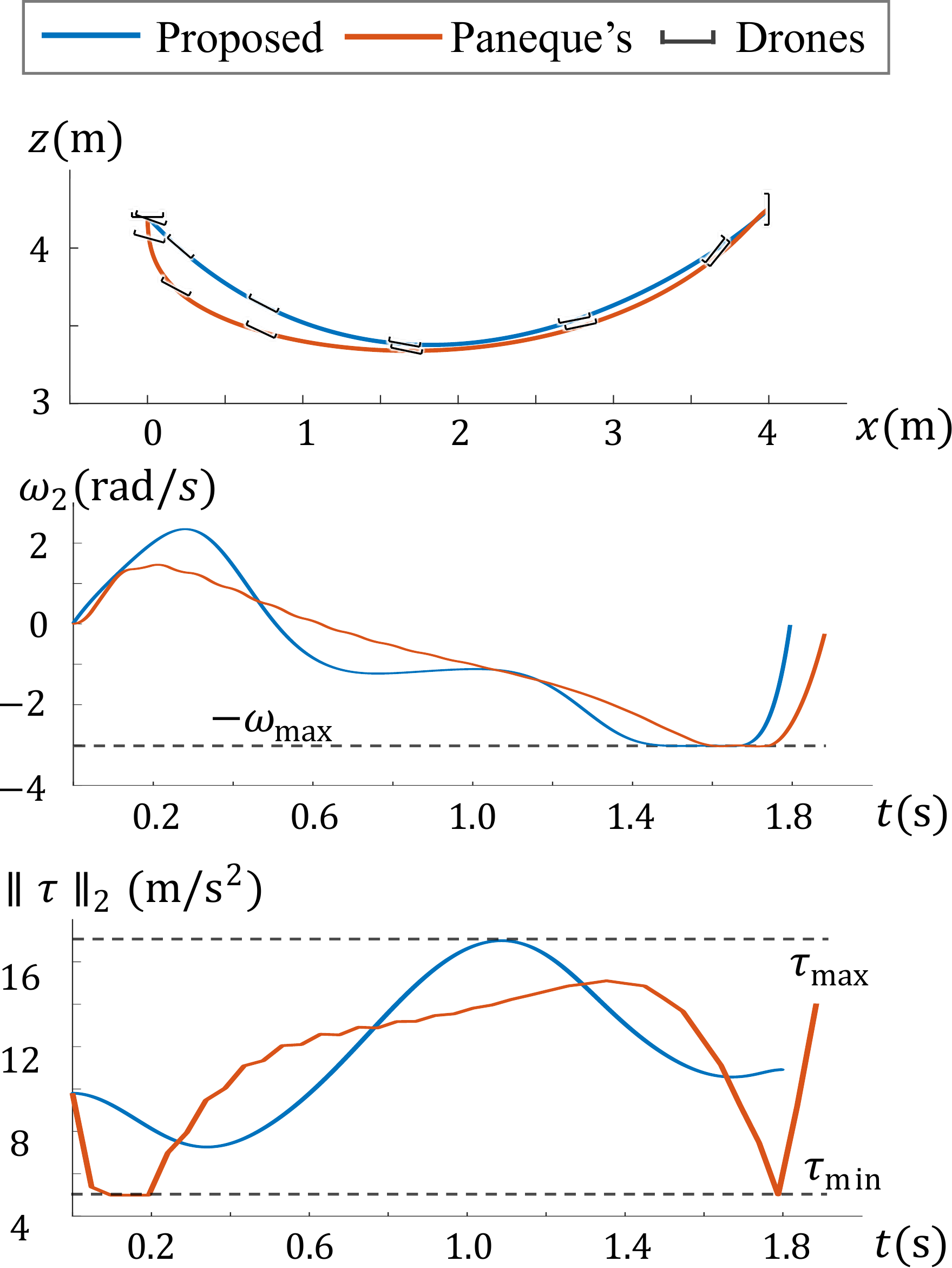}
	\end{center}
	\caption{
		\label{fig:benchmark}
		Comparison of the trajectories generated by Paneuqe's planner and ours for perching on verticle surfaces.
	}
	\vspace{-1.5cm}
\end{figure}

For our planner, we set $N = 10, \kappa = 16$ which means the constraints are imposed on 160 small segments. We set $N = 40$ for Paneque's method, which is more sparse respectively.
Taking the case of vertical surface as an example, the comparison of trajectories generated by the two methods is shown in Fig. \ref{fig:benchmark}. As we can see, the duration and the positions of the trajectories are relatively similar. Since the discrete-time NLP has more degrees of freedom, its control inputs can be changed rapidly and it needs less control effort. Nevertheless, due to the finite discrete resolution, the thrusts and body rates are less smooth than ours.

We also evaluate the computation time of the two methods for perching on inclined surfaces of different slopes.
All the simulation experiments are run on a desktop equipped with an Intel Core i5-12600 CPU. 
As is shown in Tab. \ref{tab:computation time}, the proposed method needs a much lower computation budget.

\begin{figure*}[ht]
	\begin{center}
		\includegraphics[width=2.0\columnwidth]{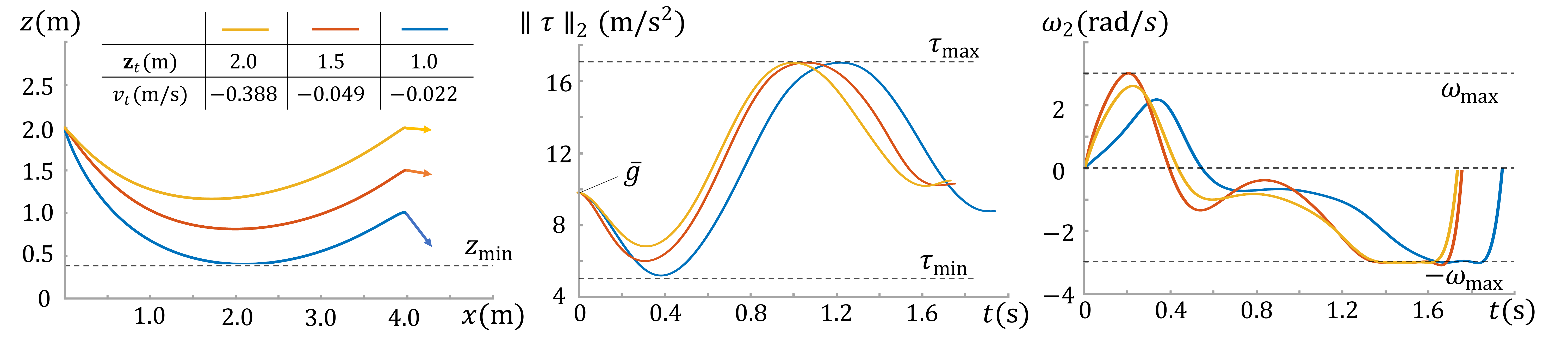}
	\end{center}
	\caption{
		\label{fig:terminal_v3}
		Simulation for testing terminal states adjustment in different cases.
	}
\end{figure*}

\begin{table}[ht]
	\renewcommand\arraystretch{1.3}
	\centering
	\caption{Computation time Comparison}
	\label{tab:computation time}
	\begin{tabular}{ccccc}
		\toprule
		\multicolumn{2}{c}{\multirow{2}{*}{Methods}} & \multicolumn{3}{c}{Average calculating time (ms)} \\
		\cmidrule(lr){3-5} 
		\multicolumn{2}{c}{}                          & $-70^\circ$     & $-90^\circ$       & $-110^\circ$           \\
		\midrule
		\multicolumn{2}{c}{Our Proposed}                      & 1.47251         & 4.60674        & 17.7263        \\
		\multicolumn{2}{c}{Paneque's}      & 257.533         & 101.450       & 81.520        \\
		\bottomrule
	\end{tabular}
\end{table}

Moreover, we conduct several simulations to test the adjustment of terminal state in different cases, shown in Fig. \ref{fig:terminal_v3}.
As the height of the landing position $\mathbf z_t$ decreases from $2.0 \mathrm m$ to $1.0 \mathrm m$, the generated trajectory has increasing tangential velocity $v_t$ to guarantee both safety and feasibility.

\section{conclusion and future work}

In this paper, we propose a highly-versatile trajectory planning framework for aerial perching on moving inclined surfaces.
Our method considers $\mathrm {SE(3)}$ motion planning for dynamic collision avoidance and is able to adjust terminal state adaptively when there is no enough space.
Sufficient experiments and benchmarks validate the robustness of our planner.
Since there is a singularity of Hopf fibration when the drone is upside down, our method cannot be directly used for this corner case. 
In the future, we will improve our method to solve this problem.
Besides, we will extend our method to autonomous perching where the estimation, target detection are obtained totally from on-board sensors.

\bibliographystyle{IEEEtran}
\bibliography{IROS2022jjl}
\end{document}